\documentclass{article}

\PassOptionsToPackage{numbers}{natbib}
\usepackage[final]{ai_2016}

\usepackage[utf8]{inputenc} 
\usepackage[T1]{fontenc}    
\usepackage{url}            
\usepackage{booktabs}       
\usepackage{amsfonts}       
\usepackage{nicefrac}       
\usepackage{microtype}      

\usepackage{todonotes}
\usepackage{mathtools}
\usepackage{amssymb}
\usepackage{amsmath}
\usepackage{algorithm}
\usepackage{algorithmicx} 
\usepackage{algpseudocode}
\usepackage{caption}
\usepackage{subcaption}
\usepackage{listings}
\usepackage{placeins}
\usepackage{framed}
\usepackage[]{ntheorem}
\usepackage{enumitem}

\usepackage{adjustbox}
\usepackage{etoolbox}

\makeatletter
\expandafter\patchcmd\csname\string\algorithmic\endcsname%
      {\labelwidth 0.5em}{\labelwidth0pt\labelsep0pt}{}{}

\algnewcommand{\LineComment}[1]{\State \(\triangleright\) #1}
\algnewcommand{\LineCommentx}[1]{\Statex \(\triangleright\) #1}

\makeatletter
\newcommand{\subalign}[1]{%
  \vcenter{%
    \Let@ \restore@math@cr \default@tag
    \baselineskip\fontdimen10 \scriptfont\tw@
    \advance\baselineskip\fontdimen12 \scriptfont\tw@
    \lineskip\thr@@\fontdimen8 \scriptfont\thr@@
    \lineskiplimit\lineskip
    \ialign{\hfil$\m@th\scriptstyle##$&$\m@th\scriptstyle{}##$\crcr
      #1\crcr
    }%
  }
}
\makeatother

\newtheorem{frm-def}{Definition}
\newtheorem{frm-prop}{Proposition}
\newtheorem{frm-lemma}{Lemma}
\newtheorem{frm-def-sup}{Definition}
\newtheorem{frm-prop-sup}{Proposition}
\newtheorem{frm-lemma-sup}{Lemma}

\title{Measuring the non-asymptotic convergence of sequential Monte Carlo samplers using probabilistic programming}

\author{
  Marco F.~Cusumano-Towner\\
  \small Computer Science \& Artificial Intelligence Laboratory\\
  \small Massachusetts Institute of Technology\\
  \texttt{marcoct@mit.edu} \\
  \And
  Vikash K.~Mansinghka\\
  \small Department of Brain \& Cognitive Sciences\\
  \small Massachusetts Institute of Technology\\
  \texttt{vkm@mit.edu} \\
}

\begin{document}

\maketitle
\vfill
\section{Introduction}
A key limitation of sampling algorithms for approximate inference is that it is difficult to quantify their approximation error. Widely used sampling schemes, such as sequential importance sampling with resampling and Metropolis-Hastings, produce output samples drawn from a distribution that may be far from the target posterior distribution. This paper shows how to upper-bound the symmetric KL divergence between the output distribution of a broad class of sequential Monte Carlo (SMC) samplers and their target posterior distributions, subject to assumptions about the accuracy of a separate gold-standard sampler. The proposed method applies to samplers that combine multiple particles, multinomial resampling, and rejuvenation kernels. The experiments show the technique being used to estimate bounds on the divergence of SMC samplers for posterior inference in a Bayesian linear regression model and a Dirichlet process mixture model.

This paper builds on a growing body of work begun by \cite{grosse2014model} and \cite{grosse2015sandwiching} into estimating upper bounds on KL divergences between a sampler's output distribution and the posterior. In variational inference, the KL divergence of the variational approximation is the gap between the variational lower bound and the log-evidence. \cite{grosse2014model} and \cite{grosse2015sandwiching} recognized that certain stochastic inference Markov chains including annealed importance sampling (AIS) and single-particle SMC can be treated as variational approximations over an extended space that includes auxiliary random choices of the sampler. A similar insight was introduced independently in \cite{salimans2015}. \cite{grosse2014model} and \cite{grosse2015sandwiching} also showed how to estimate upper bounds on the log-evidence for datasets simulated from the model using generalizations of the harmonic mean estimator, and introduced the bidirectional Monte Carlo (BDMC) technique for `sandwiching’ the log-evidence between these upper bounds and variational lower bounds. A related approach for sandwiching the partition function was previously used in the statistical physics literature \cite{hunter1993}. Finally, \cite{grosse2014model} and \cite{grosse2015sandwiching} recognized that the gap between the bounds serves as an upper bound on the KL divergence of the sampler, allowing BDMC to be used for measuring sampler accuracy on simulated datasets.

Two independent papers \cite{cusumano2016quantifying} and \cite{grosse2016measuring} built on \cite{grosse2014model} and \cite{grosse2015sandwiching} to develop the technique further in different ways. Our previous paper \cite{cusumano2016quantifying} took a probabilistic programming perspective, and showed how to estimate the KL divergence bound described in \cite{grosse2014model} and \cite{grosse2015sandwiching} for general samplers using a ‘meta-inference’ sampler that generates sampler execution histories. \cite{cusumano2016quantifying} also provided meta-inference samplers for sampling importance resampling (SIR) and particle filtering without MCMC rejuvenation kernels. \cite{cusumano2016quantifying} also introduced an upper bound on the symmetric KL divergence between the sampler output and the posterior, analyzed optional use of approximate ‘reference’ samples as surrogates for exact posterior samples (prompting the label ‘subjective divergence'), and related the tightness of the bounds to the accuracy of the meta-inference sampler. A closely related but independent work \cite{grosse2016measuring} introduced Bounding
Divergences with REverse Annealing (BREAD), which uses the same upper bound on the symmetric KL divergence given in \cite{cusumano2016quantifying}, and showed how to evaluate AIS and single-particle SMC approximate inference quality using this bound. BREAD also includes a heuristic scheme, applicable to hierarchical Bayesian statistical models, for generating simulated datasets whose divergence profiles are used as proxies for divergence profiles on real-world datasets. \cite{grosse2016measuring} also integrated their technique into existing probabilistic programming platforms.

The main contribution of the current work is a meta-inference construction for generic SMC samplers \cite{del2006sequential} that is related to conditional SMC \cite{andrieu2010particle} and generalizes the existing meta-inference constructions for AIS, single-particle SMC, SIR, and particle filtering. By handling a broad class of samplers, the construction increases relevance for real world problems. The construction allows analysis of samplers that rely on MCMC rejuvenation kernels for good inference quality, while permitting use of multiple particles (instead of custom model-specific annealing schemes) to tighten the KL divergence bounds.

\section{Background on subjective divergence}
\vspace{-3mm}
We first review the subjective divergence procedure of \cite{cusumano2016quantifying}.
Let $p$ denote an approximate inference sampling program that samples output $z \sim p(z)$ for $z \in \mathcal{Z}$. Suppose $p$ also comes endowed with a side-procedure that evaluates the log probability $\log p(z)$ that the sampler produces any given output $z$. Let $\pi(z)$ denote the posterior distribution, and let $\tilde{\pi}(z) = \pi(z) Z_{\tilde{\pi}}$ denote an unnormalized posterior distribution. Suppose that we have access to samples from $\pi(z)$. Then the following is an unbiased Monte Carlo estimate of the symmetric KL divergence between $p(z)$ and $\pi(z)$:
\begin{equation}
\frac{1}{N} \sum_{i=1}^N \log \frac{\tilde{\pi}(z_1^i)}{p(z_1^i)}
-
\frac{1}{M} \sum_{j=1}^M \log \frac{\tilde{\pi}(z_2^j)}{p(z_2^j)}
\;\;\;
\mbox{for}
\left. \begin{array}{ll}
&z_1^i \sim \pi(z)
\;\;
i=1\ldots N\\&z^j_2 \sim p(z)
\;\;
j=1\ldots M
\end{array} \right.
\end{equation}
Unfortunately, it is often not possible to efficiently evaluate $\log p(z)$ for sampling programs that sample auxiliary random choices during their execution, including MCMC and SMC sampling algorithms for approximate Bayesian inference. We denote the joint distribution over auxiliary random choices $u$ and output $z$ by $p(u,z)$. It is intractable to marginalize out the auxiliary random choices $u$ because there is an exponentially large number of terms in the sum $p(z) = \sum_u p(u,z)$. Therefore, we instead compute the following unbiased estimate of an upper bound on the symmetric KL divergence, using a `meta-inference' sampler program $u|z \sim q(u;z)$ which samples execution histories of the sampler $p$ (assignments to the auxiliary variables $u$) given the output $z$:
\begin{equation} \label{eq:subjective_divergence}
\frac{1}{N} \sum_{i=1}^N \log \frac{\tilde{\pi}(z_1^i) q(u_1^i; z_1^i)}{p(u_1^i, z_1^i)}
-
\frac{1}{M} \sum_{j=1}^M \log \frac{\tilde{\pi}(z_2^j) q(u_2^j; z_2^j)}{p(u_2^j, z_2^j)}
\;\;\;
\mbox{for}  \left. \begin{array}{ll}
&u_1^i, z_1^i \sim \pi(z) q(u;z)
\;\;
i=1\ldots N\\&u_2^j, z^j_2 \sim p(u,z)
\;\;
j=1\ldots M
\end{array}\right.
\end{equation}
The upper bound estimated is the symmetric KL divergence on an extended space that includes the auxiliary variables $u$ of the sampler. As shown in \cite{cusumano2016quantifying}, the tightness of the bound is governed by how well $q(u;z)$ approximates $p(u|z)$ on average for $z \sim p(z)$ and $z \sim \pi(z)$. When samples from a gold-standard approximate inference `reference sampler' are used in place of posterior samples, the validity of the bound is subject to the accuracy of the reference sampler \cite{cusumano2016quantifying}.
\section{A probabilistic programming interface for subjective divergence}
\vspace{-2mm}
We now clarify the procedures associated with a sampler that are needed for subjective divergence estimation. In particular, we introduce the following probabilistic programming interface, which consists of two stochastic procedures, denoted $(p,q).\textproc{simulate}$ and $(p,q).\textproc{regenerate}$ for some distributions $p(u, z)$ and $q(u;z)$:
\begingroup
\setlength{\abovedisplayskip}{6pt}
\setlength{\belowdisplayskip}{6pt}
\begin{equation}
\def\arraystretch{1.4}
\arraycolsep=0.6pt
\begin{array}{rl}
\left(z, \log (p(u,z) / q(u;z))\right) &\gets (p,q).\textproc{simulate}() \mbox{ for } u, z \sim p(u,z)  \\
     \log (p(u,z) / q(u;z)) &\gets (p,q).\textproc{regenerate}(z) \mbox{ for } u | z \sim q(u;z)
\end{array}
\end{equation}
\endgroup
The \textproc{simulate} procedure runs a sampler with joint distribution $p(u,z)$ over execution histories $u$ and output $z$, and returns $z$. The \textproc{regenerate} procedure takes a potential sampler output $z$ as its input, and runs a `regeneration' sampler that samples an execution history $u$ of the original sampler. Both procedures also return a log-weight. The log-weight returned by \textproc{simulate} can be interpreted as a log harmonic mean estimate of $p(z)$ and the log-weight returned by \textproc{regenerate} can be interpreted as a log importance sampling estimate of $p(z)$. When the sampler is an inference sampler, we call the regeneration sampler a `meta-inference' sampler. As will be seen, the relationship between the original sampler and the regeneration sampler is analogous to the relationship between SMC and conditional SMC \cite{andrieu2010particle}.

Note that the auxiliary random variables $u$ are not exposed through the interface. Also note that a sampler with a tractable marginal output probability $p(z)$ trivially implements the interface because $\log (p(u,z)/q(u;z))$ reduces to the log output probability when there are no auxiliary variables $u$. Algorithm~\ref{alg:subjective_divergence} shows a procedure that computes Equation~(\ref{eq:subjective_divergence}) using the above interface.
  \begin{algorithm}[H]
        \algrenewcommand\algorithmicindent{1em}%
        \caption{Subjective divergence estimation using \textproc{simulate} and \textproc{regenerate}} \label{alg:subjective_divergence}
        \begin{algorithmic}[0]
        \Require Sampler package $(p,q)$ implementing \textproc{simulate} and \textproc{regenerate}; posterior sampler $z \sim \pi(z)$ or reference sampler $z \sim r(z)$; unnormalized posterior probability function $\tilde{\pi}(z)$.
        \Procedure{estimate-kl-bound}{$(p,q)$, $\pi$, $\tilde{\pi}$}
        \For{$i \gets 1 \ldots N$}
            \State $z^i_1 \sim \pi(z)$ \small\Comment{Replace with sample from reference sampler $z^i_1 \sim r(z)$ if exact posterior sampler unavailable}\normalsize
            \State $\ell^i_1 \gets (p,q).\textproc{regenerate}(z^i_1)$
        \EndFor
        \For{$j \gets 1 \ldots M$}
            \State $(z_2^j, \ell^j_2) \gets (p,q).\textproc{simulate}()$        
        \EndFor        
        \State \Return $\frac{1}{N} \sum_{i=1}^N (\log \tilde{\pi}(z_1^i) - \ell^i_1) - \frac{1}{M} \sum_{j=1}^M (\log \tilde{\pi}(z^j_2) - \ell^j_2) $
        \EndProcedure
        \end{algorithmic}
  \end{algorithm}
\clearpage
\section{Implementing \textproc{simulate} and \textproc{regenerate} for sequential Monte Carlo}
\vspace{-2mm}
Algorithm~\ref{alg:smc} below shows how to implement $\textproc{simulate}$ and $\textproc{regenerate}$ for the generic SMC sampler template introduced in \cite{del2006sequential}, with independent resampling. The SMC sampler template (the \textproc{simulate} procedure of Algorithm~\ref{alg:smc}), permits use of MCMC kernels (within the $k_t$), provided that corresponding `backward kernels' $\ell_t$ are defined such that the weights can be computed. Note that $\textproc{simulate}$ does not sample from the backward kernels.
Building on the analysis of SMC used in \cite{andrieu2010particle}, the auxiliary variables $u$ for the SMC sampler are the random choices made during its execution: the resampling choices $a_t^i \in \{1\ldots N\}$ for $(i,t) \in \{1\ldots N\} \times \{1\ldots T-1\}$ and $I_T \in \{1\ldots N\}$ and the values of all intermediate particles $x^i_t \in \mathcal{X}_t$ for $(i, t) \in \{1\ldots N\} \times \{1\ldots T\}$. The output of the SMC sampler is denoted $z \in \mathcal{X}_T$. The SMC stochastic regeneration template (the \textproc{regenerate} procedure of Algorithm~\ref{alg:smc}), is given an output $z \in \mathcal{X}_T$, and samples an execution history $u$ of the SMC sampler by first choosing the ancestral particle indices that led to the output (denoted $I_t$ for $t \in \{1\ldots T\}$), then \emph{sampling} from the backward kernels $\ell_t$ in reverse order to define the ancestral particle values $x^{I_t}_t$ for $t \in \{1\ldots T\}$ that led to the output, and finally running SMC forward, with the ancestral indices $I_t$ and values $x_t^{I_t}$ for $t \in \{1\ldots T\}$ fixed. This is related to the conditional SMC update of \cite{andrieu2010particle}, but differs in that only an output particle and not a full particle trajectory is required as input. The log-weight for this sampler and regeneration pair simplify to (see Appendix~A for derivation):
\vspace{-1mm}
\begin{equation}
\log \frac{p(u,z)}{q(u;z)} = -\log \left(w_{T+1}^1 \prod_{t=1}^T \frac{1}{N} \sum_{j=1}^N w_t^j\right)
\end{equation}
\vspace{-4mm}
\begin{algorithm}
\caption{\textproc{simulate} and \textproc{regenerate} for SMC samplers with independent resampling} \label{alg:smc}
\begin{algorithmic}[0]
\Require Number of steps $T$; hypothesis spaces $\mathcal{X}_t$ (not necessarily related) and unnormalized target distributions $\tilde{p}_t$ defined on $\mathcal{X}_t$ where $\tilde{p}_t(x_t) > 0$ for $x_t \in \mathcal{X}_t$ for $t \in \{1\ldots T\}$; sampler for initialization kernel $k_1$ defined on $\mathcal{X}_1$ with $k_1(x_1) > 0$ for $x_1 \in \mathcal{X}_1$; samplers for kernels $k_t$ indexed by $\mathcal{X}_{t-1}$ and defined on $\mathcal{X}_t$ for $t \in \{2\ldots T\}$; sampler for kernel $k_{T+1}$ indexed by $\mathcal{X}_T$ and defined on $\mathcal{X}_T$; samplers for kernels $\ell_t$ indexed by $\mathcal{X}_t$ and defined on $\mathcal{X}_{t-1}$ such that $k_t(x_t;x_{t-1}) > 0 \iff \ell_t(x_{t-1};x_t) > 0$ for $x_{t-1} \in \mathcal{X}_{t-1}, x_t \in \mathcal{X}_t$ for $t \in \{2\ldots T\}$; sampler for kernel $\ell_{T+1}$ such that $k_{T+1}(z';z) > 0 \iff \ell_{T+1}(z;z') > 0$ for $z,z' \in \mathcal{X}_T$;
evaluator procedures for weight functions $w_1(x_1) := \frac{\tilde{p}_1(x_1)}{k_1(x_1)}$, $w_t(x_{t-1},x_t) := \frac{\tilde{p}_t(x_t) \ell_t(x_{t-1};x_t)}{\tilde{p}_{t-1}(x_{t-1}) k_t(x_t;x_{t-1})}$ for $t \in \{2,\ldots,T\}$ and $w_{T+1}(x_T,x'_T) := \frac{\ell_{T+1}(x_T;x'_T)}{\tilde{p}_T(x_T) k_{T+1}(x'_T;x_T)}$ for $x_T,x'_T \in \mathcal{X}_T$; number of particles $N$
\end{algorithmic}
\begin{minipage}[t]{7.5cm}
\begin{algorithmic}
\Procedure{simulate}{~}
\For{$i \gets 1\ldots N$}
    \State $x^i_1 \sim k_1(\cdot)$
    \State $w^i_1 \gets w_1(x^i_1)$
\EndFor
\For{$t \gets 2\ldots T$}
    \For{$i \gets 1\ldots N$}
        \State $a_{t-1}^i \sim \small\mbox{Categorical}(\textproc{normalize}(\mathbf{w}_{t-1}))$\normalsize
        \State $x^i_t \sim k_t(\cdot;x_{t-1}^{a_{t-1}^i})$
        \State $w_t^i \gets w_t(x_{t-1}^{a_{t-1}^i}, x^i_t)$
    \EndFor
\EndFor
\State $I_T \sim \small\mbox{Categorical}(\textproc{normalize}(\mathbf{w}_T))$\normalsize
\State $z \sim k_{T+1}(\cdot; x_T^{I_T})$
\State $w_{T+1}^1 \gets w_{T+1}(x_T^{I_T},z)$
\State \Return $\left(z, -\log \left(w_{T+1}^1 \prod_{t=1}^T \frac{1}{N} \sum_{j=1}^N w_t^j\right)\right)$ 
\EndProcedure
\State 
\Procedure{rand-ancestry}{$N$, $T$}
\For{$t \gets 1\ldots T$}
    \State $I_t \sim \mbox{Uniform}(1,\ldots,N)$
\EndFor
\State \Return $(I_1,\ldots,I_T)$
\EndProcedure
\end{algorithmic}
\end{minipage}%
\begin{minipage}[t]{8cm}
\begin{algorithmic}
\Procedure{regenerate}{$z$}
\State $(I_1, \ldots, I_T) \sim $\Call{rand-ancestry}{$N$, $T$}
\State $x_T^{I_T} \sim \ell_{T+1}(\cdot; z)$
\For{$t \gets T-1\ldots 1$}
    \State $x_t^{I_t} \sim \ell_{t+1}(\cdot; x_{t+1}^{I_{t+1}})$
\EndFor
\For{$i \gets 1\ldots N$}
    \If{$i \ne I_1$}
        \State $x^i_1 \sim k_1(\cdot)$
    \EndIf
    \State $w^i_1 \gets w_1(x^i_1)$
\EndFor
\For{$t \gets 2\ldots T$}
    \For{$i \gets 1\ldots N$}
        \If{$i = I_t$}
            \State $a_{t-1}^i \gets I_{t-1}$
        \Else
            \State $a_{t-1}^i \sim \small\mbox{Categorical}(\textproc{normalize}(\mathbf{w}_{t-1}))$\normalsize
            \State $x^i_t \sim k_t(\cdot;x_{t-1}^{a_{t-1}^i})$
        \EndIf
        \State $w_t^i \gets w_t(x_{t-1}^{a_{t-1}^i}, x^i_t)$
    \EndFor
\EndFor
\State $w_{T+1}^1 \gets w_{T+1}(x_T^{I_T},z)$
\State \Return $-\log \left(w_{T+1}^1 \prod_{t=1}^T \frac{1}{N} \sum_{j=1}^N w_t^j\right)$
\EndProcedure
\end{algorithmic}
\end{minipage}
\end{algorithm}

\begin{figure}[h]
    \centering
    \begin{minipage}{1.0\textwidth}
        \centering
        \begin{subfigure}[b]{0.5\textwidth}
            \centering
            \includegraphics[width=1.0\textwidth]{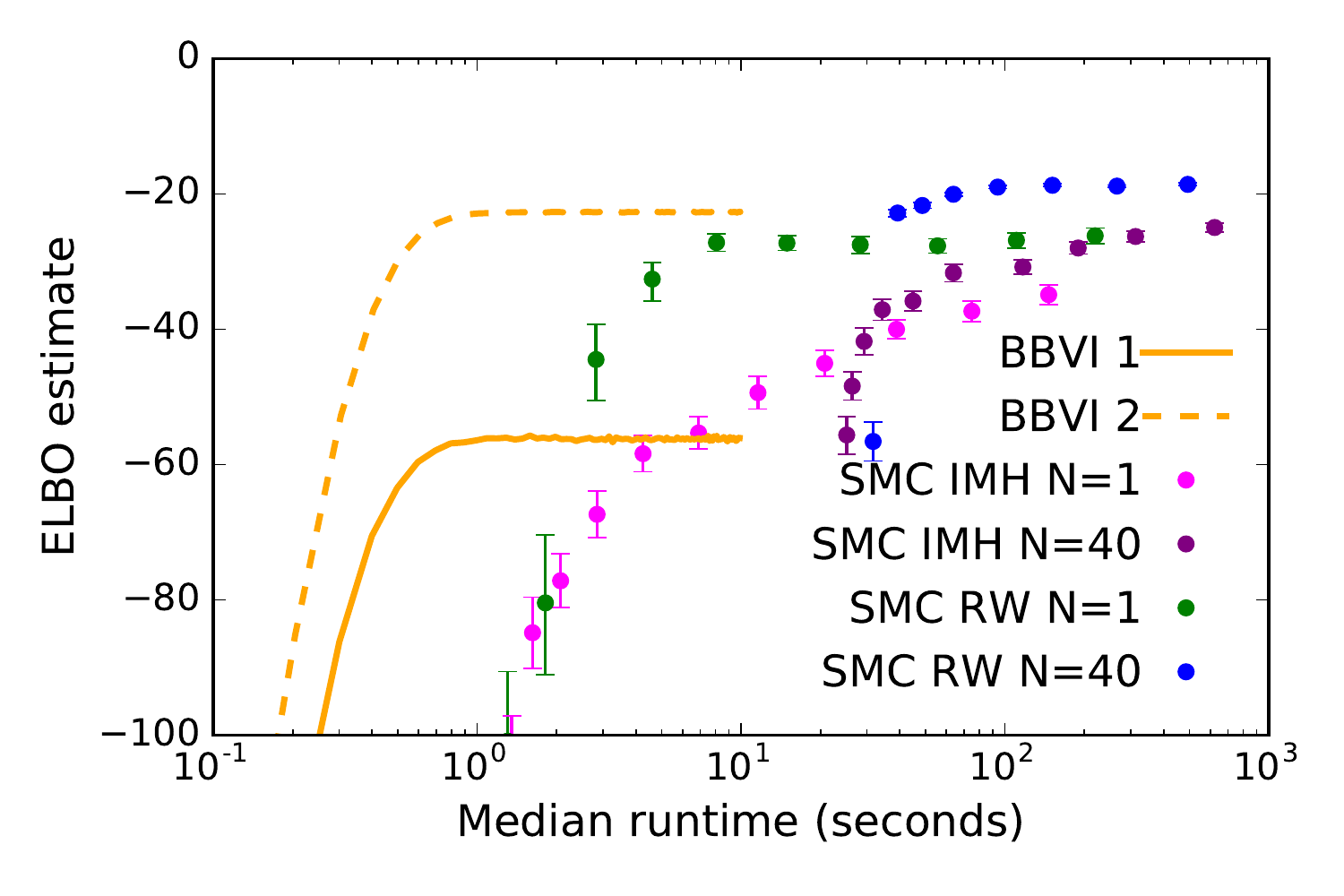}
            \caption{~}
            \label{fig:linreg_elbo}
        \end{subfigure}%
        \begin{subfigure}[b]{0.5\textwidth}
            \centering
            \includegraphics[width=1.0\textwidth]{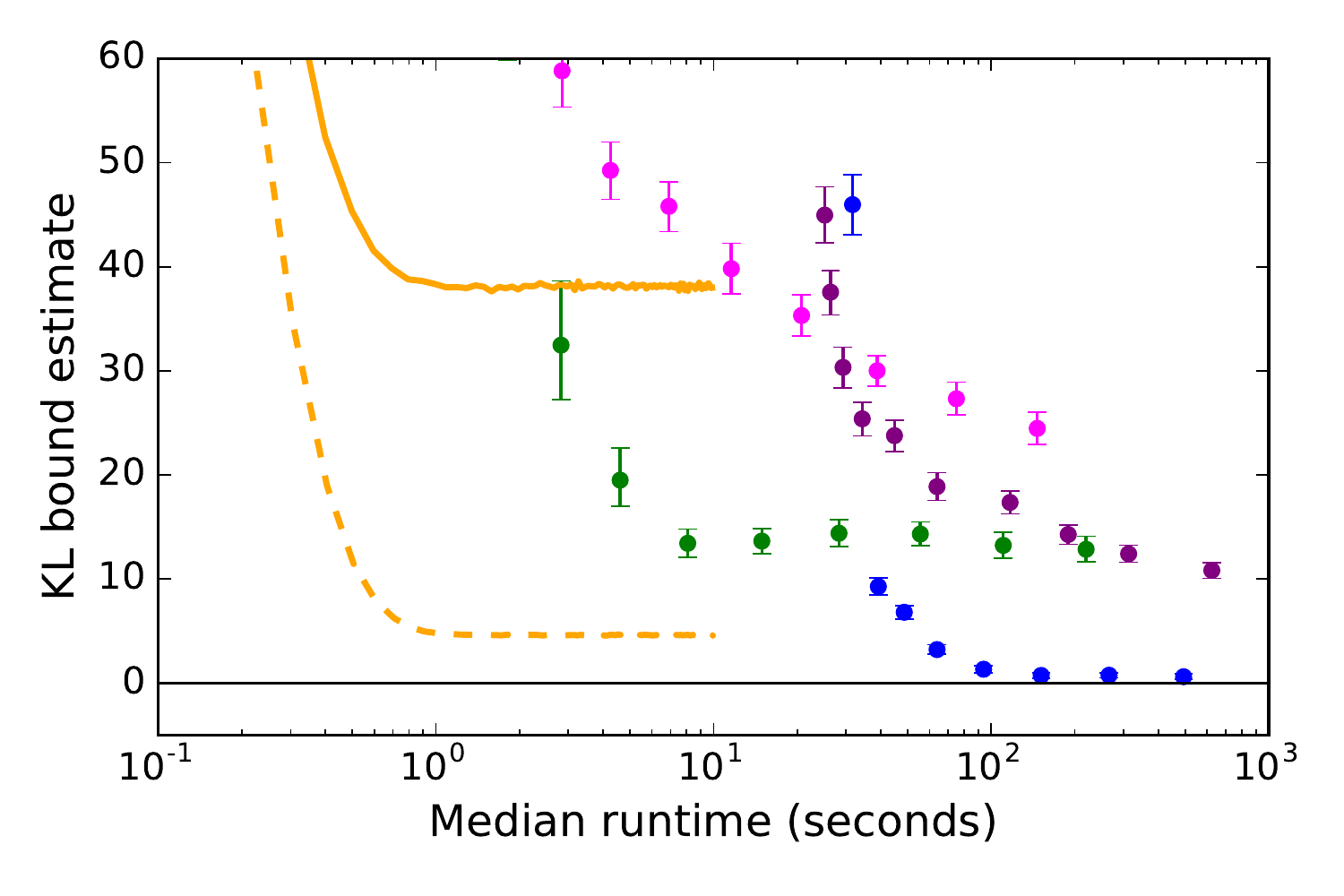}
            \caption{~}
            \label{fig:linreg_sd}
        \end{subfigure}\\
        \begin{subfigure}[b]{0.5\textwidth}
            \centering
            \includegraphics[width=1.0\textwidth]{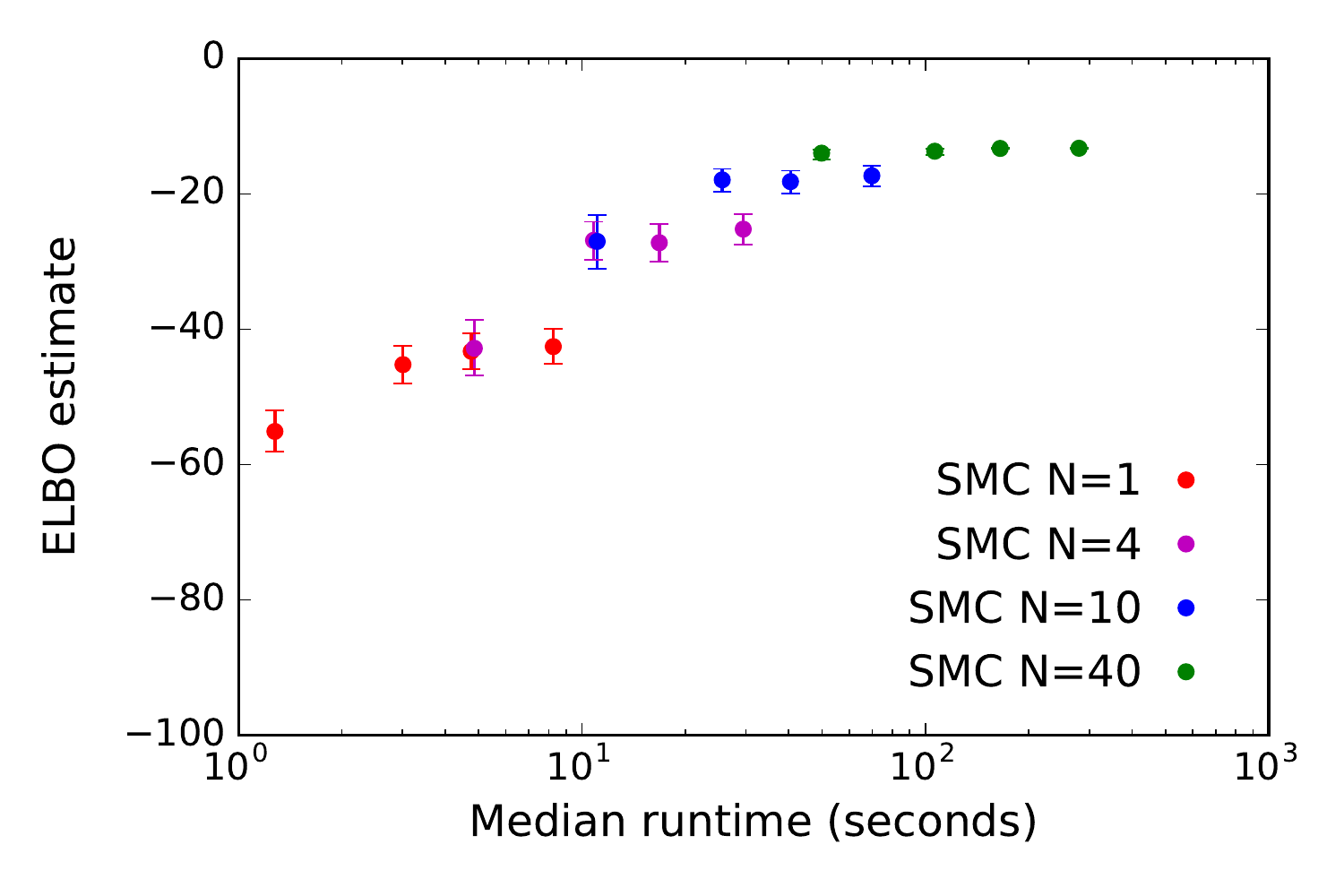}
            \caption{~}
            \label{fig:dpm_elbo}
        \end{subfigure}%
        \begin{subfigure}[b]{0.5\textwidth}
            \centering
            \includegraphics[width=1.0\textwidth]{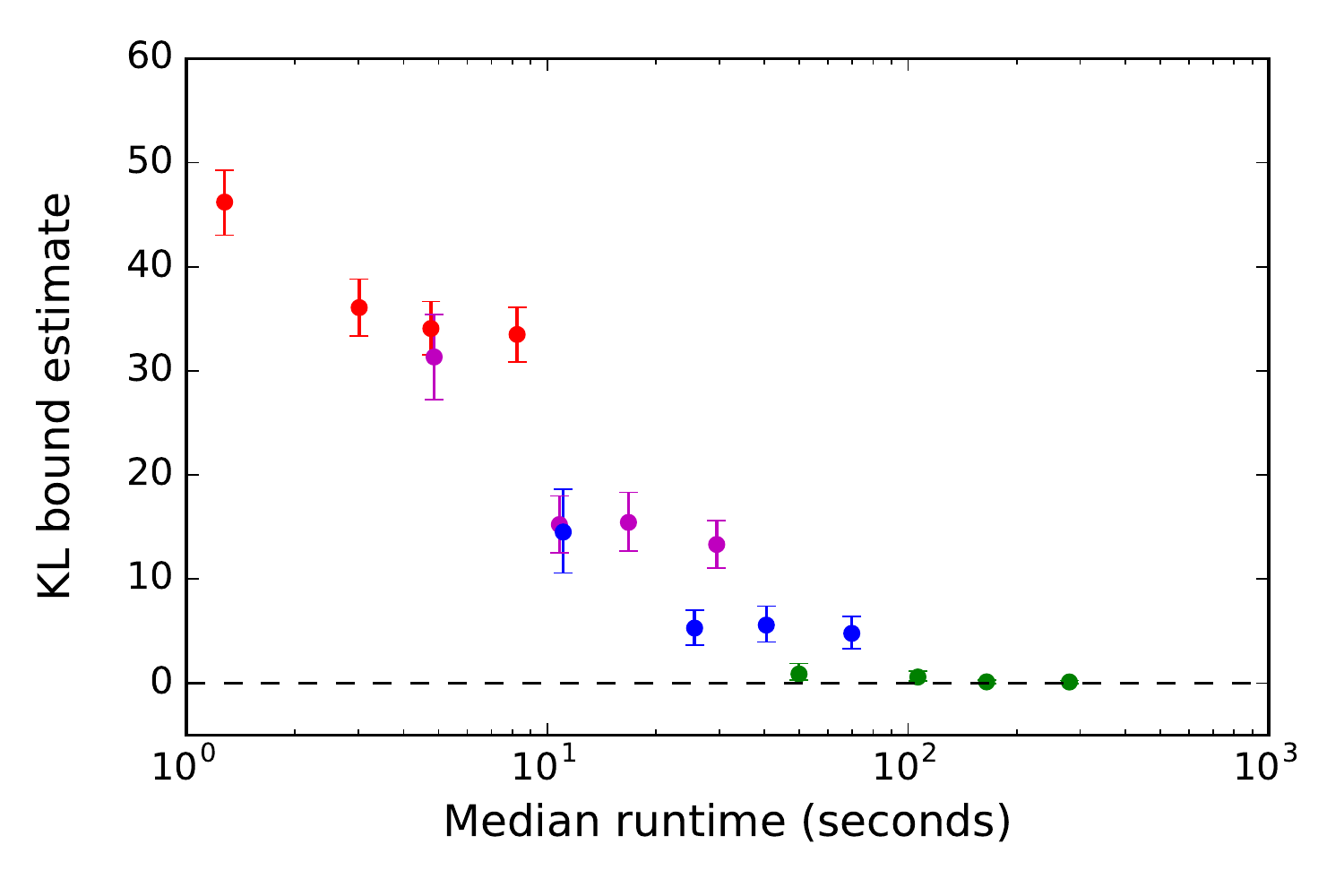}
            \caption{~}
            \label{fig:dpm_sd}
        \end{subfigure}
    \end{minipage}
    \caption{(a) and (b) show estimated ELBO lower bounds and estimated upper bounds on KL divergence to the posterior, respectively, for SMC samplers and two black box variational inference (BBVI) programs, in a Bayesian linear regression inference problem. SMC IMH samplers use single-site independent Metropolis-Hastings (MH) rejuvenation kernels, and SMC RW use single-site random-walk MH rejuvenation kernels. BBVI 1 and BBVI 2 optimize over different variational families. (c) and (d) show estimated ELBO lower bounds and subjective upper bounds on KL divergence, for SMC with single-site Gibbs rejuvenation kernels over cluster assignments in a Dirichlet process mixture model problem with collapsed cluster parameters. The SMC samplers in both problems use sequential observation to define the sequence of target distributions, and were parameterized by the number of particles (N, represented by color) and the number of applications of the MCMC rejuvenation kernels between target distribution updates (distinct estimates of the same color). Particles were initialized from the prior. Exact posterior reference samples were used for KL bound estimation in (a) and (b), and samples from a gold-standard approximate MCMC reference sampler were used in lieu of posterior samples for KL bound estimation in (c) and (d). In (a) and (b), the random-walk MH kernels appear more effective than the independent MH kernels. Increasing the number of particles tightens the KL divergence bound when the effect of rejuvenation kernels has already been saturated (compare SMC RW N=1 to SMC RW N=40). }
    \label{fig:results}
\end{figure}

\clearpage
Having specified how to implement \textproc{simulate} and \textproc{regenerate} for this generic variant of SMC, we can now estimate subjective divergences for SMC. We illustrate the use of Algorithm~\ref{alg:subjective_divergence} and Algorithm~\ref{alg:smc} to estimate subjective bounds on symmetric KL divergences of SMC samplers and black box variational approximations to the posterior in Figure~\ref{fig:results}. Note that we optimized the performance of variational inference and SMC implementations separately, and the relative runtimes of the two approaches are not meant to be informative.

\subsubsection*{Acknowledgements}
This research was supported by DARPA (PPAML program, contract number
FA8750-14-2-0004), IARPA (under research contract 2015-15061000003), the Office
of Naval Research (under  research  contract N000141310333), the Army Research
Office (under agreement number W911NF-13-1-0212),  and gifts from Analog
Devices and Google.
This research was conducted with Government support under and awarded by DoD, Air Force Office of Scientific Research, National Defense Science and Engineering Graduate (NDSEG) Fellowship, 32 CFR 168a.

\begingroup
    \setlength{\bibsep}{4pt}
    \bibliographystyle{unsrt}
    \bibliography{ai_2016_references} 
\endgroup

\clearpage
\appendix
\section*{Appendix A: Derivation of log weight for SMC}
Recall that the set of auxiliary random choices $u$ for the SMC sampler of Algorithm~\ref{alg:smc} is the set of all resampling choices $a_t^i \in \{1\ldots N\}$ for $(i,t) \in \{1\ldots N\} \times \{1\ldots T-1\}$ and $I_T \in \{1\ldots N\}$ and the values of all intermediate particles $x^i_t \in \mathcal{X}_t$ for $(i, t) \in \{1\ldots N\} \times \{1\ldots T\}$.
The joint probability over auxiliary random choices $u$ and output $z$ for an execution of SMC's \textproc{simulate} is:
\begin{align*}
p(u, z) := \left[ \prod_{i=1}^N k_1(x^i_1) \right] \left[ \prod_{t=2}^T \prod_{i=1}^N \frac{w_{t-1}^{a^i_{t-1}}}{\sum_{j=1}^N w_{t-1}^j} k_t(x^i_t; x_{t-1}^{a_{t-1}^i})\right]
\left[ \frac{w_T^{I_T}}{\sum_{j=1}^N w_T^j} k_{T+1}(z;x_T^{I_T})\right]
\end{align*}
The joint probability over auxiliary random choices $u$ for an execution of SMC's \textproc{regenerate} is:
\begin{align*}
q(u; z) := \left[ \frac{1}{N^T} \right] \left[ \ell_{T+1}(x_T^{I_T}; z) \prod_{t=2}^T \ell_t(x_{t-1}^{I_{t-1}}; x_t^{I_t}) \right] \left[ \prod_{\substack{i=1\\i \ne I_1}}^N k_1(x^i_1) \right] \left[ \prod_{t=2}^T \prod_{\substack{i=1\\i \ne I_t}}^N \frac{w_{t-1}^{a^i_{t-1}}}{\sum_{j=1}^N w_{t-1}^j} k_t(x^i_t; x_{t-1}^{a_{t-1}^i}) \right]
\end{align*}
where the first $N^{-T}$ factor is due to \textproc{rand-ancestry}.
First, note that $w_t^i > 0$ for all $(i, t) \in \{1\ldots N\} \times \{1\ldots T\}$ for $u, z \sim p(u, z)$ and for $u | z \sim q(u;z)$. This is true for $t = 1$ by the requirements $\tilde{p}_1(x_1) > 0$ and $k_1(x_1) > 0$ for all $x_1 \in \mathcal{X}_1$. For $t \in \{2\ldots T\}$, $w_t^i = \frac{\tilde{p}_t(x_t^i) \ell_t(x_{t-1}^{a_{t-1}^i}; x_t^i)}{\tilde{p}_{t-1}(x_{t-1}^{a_{t-1}^i}) k_t(x_t^i; x_{t-1}^{a_{t-1}^i})}$. Either $x_t^i \sim k_t(\cdot; x_{t-1}^{a_{t-1}^i})$ or $x_{t-1}^{a_{t-1}^i} \sim \ell_t(\cdot; x_t^i)$. Using the requirements $\tilde{p}_{t}(x_t) > 0$ for all $x_t \in \mathcal{X}_t$ for all $t \in \{1\ldots T\}$ and $k_t(x_t; x_{t-1}) > 0\iff \ell_t(x_{t-1}; x_t) > 0$ for all $x_{t-1} \in \mathcal{X}_{t-1}, x_t \in \mathcal{X}_t$ for all $t \in \{2\ldots T\}$, gives $w_t^i > 0$ for all $(i,t) \in \{1\ldots N\} \times \{2\ldots T\}$.

To see that $p(u, z) > 0 \iff q(u; z) > 0$, first consider some $u, z$ such that $p(u, z) > 0$. Since $p(u, z) > 0$ we have $k_1(x_1^i) > 0$ for $i \in \{1\ldots N\}$. We also have $k_t(x_t^i; x_{t-1}^{a_{t-1}^i}) > 0$ which implies $\ell_t(x_{t-1}^{a_{t-1}^i}; x_t^i) > 0$ for $(i, t) \in \{1\ldots N\} \times \{2\ldots T\}$, and $k_{T+1}(z; x_T^{I_T}) > 0$, which implies $\ell_{T+1}(x_T^{I_T}; z) > 0$. Combined with $w_t^i > 0$ for all $(i, t) \in \{1\ldots N\} \times \{1\ldots T\}$ these ensure $q(u; z)$ is defined for output $z$ and $q(u; z) > 0$.

Next, assume $q(u; z)$ is defined for output $z$ and $q(u; z) > 0$. Then we have $\ell_{T+1}(x_T^{I_T}; z) > 0$ which implies $k_{T+1}(z; x_T^{I_T}) > 0$. We also have $\ell_t(x_{t-1}^{I_{t-1}}; x_t^{I_t}) > 0$ for $t \in \{2\ldots T\}$, which implies $k_t(x_t^i; x_{t-1}^{a_{t-1}^i}) > 0$ for $(i, t) \in \{(I_t, t) | t \in \{2\ldots T\}\}$. We also have $k_t(x_t^i; x_{t-1}^{a_{t-1}^i}) > 0$ for $(i, t) \in \{(i, t) | i \ne I_t, t \in \{2\ldots T\}\}$. Therefore $p(u, z) > 0$.

The weight $p(u, z) / q(u; z)$ is then defined for all $u, z \sim p(u, z)$ and $u | z \sim q(u; z)$, and is:
\small
\begin{align*}
\frac{p(u,z)}{q(u;z)} &= \frac{
\left[ \prod_{i=1}^N k_1(x^i_1) \right] \left[ \prod_{t=2}^T \prod_{i=1}^N \frac{w_{t-1}^{a^i_{t-1}}}{\sum_{j=1}^N w_{t-1}^j} k_t(x^i_t; x_{t-1}^{a_{t-1}^i})\right]
\left[\frac{w_T^{I_T}}{\sum_{j=1}^N w_T^j} k_{T+1}(z;x_T^{I_T})\right]
}{
\left[ \frac{1}{N^T} \right] \left[ \ell_{T+1}(x_T^{I_T}; z) \prod_{t=2}^T \ell_t(x_{t-1}^{I_{t-1}}; x_t^{I_t}) \right] \left[ \prod_{\substack{i=1\\i \ne I_1}}^N k_1(x^i_1) \right] \left[ \prod_{t=2}^T \prod_{\substack{i=1\\i \ne I_t}}^N \frac{w_{t-1}^{a^i_{t-1}}}{\sum_{j=1}^N w_{t-1}^j} k_t(x^i_t; x_{t-1}^{a_{t-1}^i}) \right]
}\\
&= \frac{
k_1(x^{I_1}_1) \left[ \prod_{t=2}^T \prod_{i=1}^N w_{t-1}^{a^i_{t-1}} k_t(x^i_t; x_{t-1}^{a_{t-1}^i})\right]
\left[w_T^{I_T} k_{T+1}(z;x_T^{I_T})\right]
}{
\left[ \prod_{t=1}^T \frac{1}{N} \sum_{j=1}^N w_t^j \right] \left[ \ell_{T+1}(x_T^{I_T}; z) \prod_{t=2}^T \ell_t(x_{t-1}^{I_{t-1}}; x_t^{I_t}) \right] \left[ \prod_{t=2}^T \prod_{\substack{i=1\\i \ne I_t}}^N w_{t-1}^{a^i_{t-1}} k_t(x^i_t; x_{t-1}^{a_{t-1}^i}) \right]
}\\
&= \frac{
k_1(x^{I_1}_1) \left[ \prod_{t=2}^T w_{t-1}^{I_{t-1}} k_t(x^{I_t}_t; x_{t-1}^{I_{t-1}})\right]
\left[w_T^{I_T} k_{T+1}(z;x_T^{I_T})\right]
}{
\left[ \prod_{t=1}^T \frac{1}{N} \sum_{j=1}^N w_t^j \right] \left[ \ell_{T+1}(x_T^{I_T}; z) \prod_{t=2}^T \ell_t(x_{t-1}^{I_{t-1}}; x_t^{I_t}) \right]
}\\
&\;= \frac{
k_1(x^{I_1}_1) \left[ \prod_{t=2}^T \frac{k_t(x_t^{I_t}; x_{t-1}^{I_{t-1}})}{\ell_t(x_{t-1}^{I_{t-1}}; x_t^{I_t})}  \right] \frac{k_{T+1}(z;x_T^{I_T})}{\ell_{T+1}(x_T^{I_T}; z)} 
\left[ \frac{\tilde{p}_1(x^{I_1}_1)}{k_1(x^{I_1}_1)} \prod_{t=2}^T \frac{\tilde{p}_t(x_t^{I_t}) \ell_t(x_{t-1}^{I_{t-1}}; x_t^{I_t})}{\tilde{p}_{t-1}(x_{t-1}^{I_{t-1}}) k_t(x_t^{I_t}; x_{t-1}^{I_{t-1}})} \right]
}{
\left[ \prod_{t=1}^T \frac{1}{N} \sum_{j=1}^N w_t^j \right]
}\\
&\;= \frac{\tilde{p}_1(x^{I_1}_1) \left[\prod_{t=2}^T \frac{\tilde{p}_t(x_t^{I_t})}{\tilde{p}_{t-1}(x_{t-1}^{I_{t-1}})} \right] \frac{k_{T+1}(z;x_T^{I_T})}{\ell_{T+1}(x_T^{I_T}; z)} }{\prod_{t=1}^T \frac{1}{N} \sum_{j=1}^N w_t^j}
= \frac{\tilde{p}_T(x_T^{I_T}) \frac{k_{T+1}(z;x_T^{I_T})}{\ell_{T+1}(x_T^{I_T}; z)} }{\prod_{t=1}^T \frac{1}{N} \sum_{j=1}^N w_t^j}
= \frac{1}{w_{T+1}^1\prod_{t=1}^T \frac{1}{N} \sum_{j=1}^N w_t^j}
\end{align*}
\normalsize

\newpage
\section*{Appendix B: SMC with sequential observation and detailed balance kernels}
In the experiments, we use SMC programs defined as follows. Let $\mathcal{X}_0$ be a hypothesis space, corresponding to `global' latent variables. Let $\mathcal{E}_t$ for $t \in \{1\ldots T\}$ be additional hypothesis space extensions, corresponding for `local' latent variables for each of $T$ observations $y_t \in \mathcal{Y}_t$ for $t \in \{1\ldots T\}$. Define $\mathcal{X}_t := \mathcal{X}_{t-1} \times \mathcal{E}_t$ for $t \in \{1\ldots T\}$. We use indexing notation where $a_{s:t} := (a_s, \ldots, a_t)$ for any $a,s,t$. Let $p(\theta, e_{1:T}, y_{1:T})$ denote the model's joint probability of $(\theta,e_{1:T},y_{1:T}) \in \mathcal{X}_T \times \mathcal{Y}_1 \times \cdots \times \mathcal{Y}_T$, where $\theta \in \mathcal{X}_0$ and $e_t \in \mathcal{E}_t$ for $t \in \{1\ldots T\}$. Assume for the given observation set $y_{1:T}$ that $p(\theta, e_{1:T}, y_{1:T}) > 0$ for all $(\theta, e_{1:T}) \in \mathcal{X}_T$. The target distribution of the SMC algorithm is the conditional distribution $p(\theta, e_{1:T} | y_{1:T}) \propto p(\theta, e_{1:T}, y_{1:T})$. Define the intermediate target distributions by $p_t(\theta, e_{1:t}) := p(\theta, e_{1:t} | y_{1:t})$, and the unnormalized target probability functions by $\tilde{p}_t(\theta, e_{1:t}) := p(\theta,e_{1:t},y_{1:t})$ for $t \in \{1\ldots T\}$. Define the initialization kernel as: $k_1(\theta, e_1) := p(\theta, e_1)$. This kernel samples from the model's prior distribution over the global latents and the local latents for the first observation. Suppose there exist `detailed balance kernels' $d_t(\theta', e'_{1:t}; \theta, e_{1:t})$ for $t \in \{1,\ldots,T\}$. Kernel $d_t$ is a collection of distributions over elements of $\mathcal{X}_t$, indexed by elements of $\mathcal{X}_t$. Each detailed balance kernel $d_t$ must satisfy the detailed balance property with respect to the intermediate target distribution $p_t$:
\begin{equation} \label{eq:detailed_balance}
d_t(\theta', e'_{1:t}; \theta, e_{1:t})p_t(\theta, e_{1:t}) = d_t(\theta, e_{1:t}; \theta', e'_{1:t}) p_t(\theta', e'_{1:t}) \;\; \mbox{for all} \;\; (\theta,e_{1:t}), (\theta',e'_{1:t}) \in \mathcal{X}_t
\end{equation}
Equivalently:
\begin{equation}
d_t(\theta', e'_{1:t}; \theta, e_{1:t})\tilde{p}_t(\theta, e_{1:t}) = d_t(\theta, e_{1:t}; \theta', e'_{1:t}) \tilde{p}_t(\theta', e'_{1:t}) \;\; \mbox{for all} \;\; (\theta,e_{1:t}), (\theta',e'_{1:t}) \in \mathcal{X}_t
\end{equation}
Define $k_t(\theta',e'_{1:t}; \theta,e_{1:t-1}) := d_{t-1}(\theta', e'_{1:t-1}; \theta, e_{1:t-1}) p(e'_t | \theta', e'_{1:t-1}, y_{1:t-1})$ for $t \in \{2\ldots T\}$. Each $k_t$ for $t \in \{2\ldots T\}$ is a collection of distributions over $\mathcal{X}_t$, indexed by elements of $\mathcal{X}_{t-1}$. It is possible to sample from $k_t$ by sampling from the detailed balance kernel $d_{t-1}$ and then sampling from the model prior distribution over the new local latents $p(e'_t | \theta', e'_{1:t-1}, y_{1:t-1})$. Intuitively, the kernel $k_t$ first performs inference $d_{t-1}$ targeting $p(\theta, e_{1:t-1} | y_{1:t-1})$, then extends the hypothesis space to include values of the local latent variables for observation $t$ by sampling from the prior. Define $k_{T+1} := d_T$. Intuitively, kernel $k_{T+1}$ performs inference $d_T$ targeting the final target distribution $p(\theta, e_{1:T} | y_{1:T})$. Define the `backward kernels' by $\ell_t(\theta', e'_{1:t-1}; \theta, e_{1:t}) := d_{t-1}(\theta', e'_{1:t-1}; \theta, e_{1:t-1})$ for $t \in \{2\ldots T\}$. Each kernel $\ell_t$ for $t \in \{2\ldots T\}$ is a collection of distributions over $\mathcal{X}_{t-1}$, indexed by elements of $\mathcal{X}_t$. To sample from $\ell_t$, we simply sample from the detailed balance kernel $d_{t-1}$. Finally, define $\ell_{T+1} := d_T$.
First, we show that $k_t(\theta', e'_{1:t}; \theta, e_{1:t-1}) > 0 \iff \ell_t(\theta, e_{1:t-1}; \theta', e'_{1:t}) > 0$ for all $t \in \{2\ldots T\}$. This follows from the detailed balance requirement (Equation~\ref{eq:detailed_balance}) and from the fact that $p_t(\theta, e_{1:t}) > 0$ for all $(\theta, e_{1:t}) \in \mathcal{X}_t$ and for all $t \in \{2\ldots T\}$. The same argument applies to $k_{T+1}$ and $\ell_{T+1}$. Note that we do not require the detailed balance kernels to be ergodic. For example, a given kernel $d_t$ may only update one of the components of $(\theta, e_{1:t})$.
Given these definitions, the weight functions become:
\begin{equation}
w_1(\theta, e_1) := \frac{\tilde{p}_1(\theta, e_1)}{k_1(\theta, e_1)} = \frac{p(\theta, e_1, y_1)}{p(\theta, e_1)} = p(y_1 | \theta, e_1)
\end{equation}
\begin{align}
w_t((\theta, e_{1:t-1}), (\theta', e'_{1:t})) &:= \frac{\tilde{p}_t(\theta', e'_{1:t})}{\tilde{p}_{t-1}(\theta, e_{1:t-1})} \frac{\ell_t(\theta, e_{1:t-1}; \theta', e'_{1:t})}{k_t(\theta', e'_{1:t}; \theta, e_{1:t-1})}\\
&= \frac{p(\theta', e'_{1:t}, y_{1:t})}{p(\theta, e_{1:t-1}, y_{1:t-1})} \frac{d_{t-1}(\theta, e_{1:t-1}; \theta', e'_{1:t-1})}{d_{t-1}(\theta', e'_{1:t-1}; \theta, e_{1:t-1}) p(e'_t | \theta', e'_{1:t-1}, y_{1:t-1})}\\
\end{align}
Then by detailed balance:
\begin{equation}
\frac{d_{t-1}(\theta, e_{1:t-1}; \theta', e'_{1:t-1})}{d_{t-1}(\theta', e'_{1:t-1}; \theta, e_{1:t-1})} = \frac{\tilde{p}_{t-1}(\theta, e_{1:t-1})}{\tilde{p}_{t-1}(\theta', e'_{1:t-1})} = \frac{p(\theta, e_{1:t-1}, y_{1:t-1})}{p(\theta', e'_{1:t-1}, y_{1:t-1})}
\end{equation}
\begin{align}
w_t((\theta, e_{1:t-1}), (\theta', e'_{1:t})) &= \frac{p(\theta', e'_{1:t}, y_{1:t})}{p(\theta, e_{1:t-1}, y_{1:t-1})} \frac{p(\theta, e_{1:t-1}, y_{1:t-1})}{p(\theta', e'_{1:t-1}, y_{1:t-1})p(e'_t | \theta', e'_{1:t-1}, y_{1:t-1})}\\
&= \frac{p(\theta', e'_{1:t}, y_{1:t})}{p(\theta, e_{1:t-1}, y_{1:t-1})} \frac{p(\theta, e_{1:t-1}, y_{1:t-1})}{p(\theta', e'_{1:t}, y_{1:t-1})}
= \frac{p(\theta', e'_{1:t}, y_{1:t})}{p(\theta', e'_{1:t}, y_{1:t-1})}\\
&= p(y_t | \theta', e'_{1:t}, y_{1:t-1})
\end{align}
Finally, $w_{T+1}((\theta, e_{1:T}), (\theta', e_{1:T}')) = 1/p(\theta', e_{1:T}', y_{1:T})$. Algorithm~\ref{alg:smc_experiments} shows \textproc{simulate} and \textproc{regenerate} specialized for sequential observation and detailed balance kernels, as used in the experiments. In Algorithm~\ref{alg:smc_experiments}, parenthesized superscripts indicate the step $t$ of the SMC algorithm, whereas subscripts indicate observation indices (e.g. $e_{t-1}^{i(t)}$ is the value of local latents for observation $t-1$ in particle $i$ at step $t$ of SMC).

\begin{algorithm}
\caption{\textproc{simulate} and \textproc{regenerate} for SMC with sequential observation and detailed balance} \label{alg:smc_experiments}
\begin{algorithmic}
\Procedure{simulate}{~}
\For{$i \gets 1\ldots N$}
    \State $\theta^{i(1)} \sim p(\theta)$ \Comment{Sample global latents $\theta$ from the prior}
    \State $e_1^{i(1)} \sim p(e_1 | \theta)$ \Comment{Sample local latents for observation $1$ by forward sampling in the model}
    \State $w^i_1 \gets p(y_1 | \theta^{i(1)}, e_1^{i(1)})$ \Comment{Evaluate likelihood of $y_1$}
\EndFor
\For{$t \gets 2\ldots T$}
    \For{$i \gets 1\ldots N$}
        \State $a_{t-1}^i \sim \mbox{Categorical}(\textproc{normalize}(\mathbf{w}_{t-1}))$ \Comment{Sample the index of the parent particle}
        \State $\theta^{i(t)}, e_{1:t-1}^{i(t)} \sim d_{t-1}(\cdot; \theta^{a_{t-1}^i(t-1)}, e_{1:t-1}^{a_{t-1}^i(t-1)})$ \footnotesize\Comment{Detailed balance targeting $p(\theta, e_{1:t-1} | y_{1:t-1})$}\normalsize
        \State $e_t^{i(t)} \sim p(e_t | \theta^{i(t)}, e_{1:t-1}^{i(t)})$ \small\Comment{Sample local latents for observation $t$ by forward sampling in the model}\normalsize
        \State $w_t^i \gets p(y_t | \theta^{i(t)}, e_{1:t}^{i(t)})$ \Comment{Evaluate likelihood of $y_t$}
    \EndFor
\EndFor
\State $I_T \sim \mbox{Categorical}(\textproc{normalize}(\mathbf{w}_T))$ \Comment{Sample output particle index}
\State $\theta, e_{1:T} \sim d_T(\cdot; \theta^{I_T(T)}, e_{1:T}^{I_T(T)})$ \Comment{Detailed balance targeting $p(\theta, e_{1:T} | y_{1:T})$}
\State \Return $\left(\left(\theta, e_{1:T}\right), \log \frac{p(\theta, e_{1:T}, y_{1:T})}{\prod_{t=1}^T \frac{1}{N} \sum_{j=1}^N w_t^j}\right)$ \Comment{Return output sample and log-weight}
\EndProcedure
\Statex
\Procedure{regenerate}{$\left(\theta, e_{1:T}\right)$}
\State $(I_1, \ldots, I_T) \sim $\Call{rand-ancestry}{$N$, $T$}
\State $\theta^{I_T(T)}, e_{1:T}^{I_T(T)} \sim d_T(\cdot; \theta, e_{1:T})$ \Comment{Detailed balance targeting $p(\theta, e_{1:T} | y_{1:T})$}
\For{$t \gets T-1\ldots 1$}
    \State $\theta^{I_t(t)}, e_{1:t}^{I_t(t)} \sim d_t(\cdot; \theta^{I_{t+1}(t+1)}, e_{1:t}^{I_{t+1}(t+1)})$ \Comment{Detailed balance targeting $p(\theta, e_{1:t} | y_{1:t})$}
\EndFor
\For{$i \gets 1\ldots N$}
    \If{$i \ne I_1$}
        \State $\theta^{i(1)} \sim p(\theta)$ \Comment{Sample global latents $\theta$ from the prior}
        \State $e_1^{i(1)} \sim p(e_1 | \theta)$ \Comment{Sample local latents for observation $1$ by forward sampling in the model}
    \EndIf
    \State $w^i_1 \gets p(y_1 | \theta^{i(1)}, e_1^{i(1)})$ \Comment{Evaluate likelihood of $y_1$}
\EndFor
\For{$t \gets 2\ldots T$}
    \For{$i \gets 1\ldots N$}
        \If{$i = I_t$}
            \State $a_{t-1}^i \gets I_{t-1}$
        \Else
            \State $a_{t-1}^i \sim \mbox{Categorical}(\textproc{normalize}(\mathbf{w}_{t-1}))$ \Comment{Sample the index of the parent particle}
            \State $\theta^{i(t)}, e_{1:t-1}^{i(t)} \sim d_{t-1}(\cdot; \theta^{a_{t-1}^i(t-1)}, e_{1:t-1}^{a_{t-1}^i(t-1)})$ \footnotesize\Comment{Detailed balance targeting $p(\theta, e_{1:t-1} | y_{1:t-1})$}\normalsize
            \State $e_t^{i(t)} \sim p(e_t | \theta^{i(t)}, e_{1:t-1}^{i(t)})$ \small\Comment{Sample local latents for observation $t$ by forward sampling in the model}\normalsize
        \EndIf
        \State $w_t^i \gets p(y_t | \theta^{i(t)}, e_{1:t}^{i(t)})$ \Comment{Evaluate likelihood of $y_t$}
    \EndFor
\EndFor
\State \Return $\log \frac{p(\theta, e_{1:T}, y_{1:T})}{\prod_{t=1}^T \frac{1}{N} \sum_{j=1}^N w_t^j}$ \Comment{Return the log-weight}
\EndProcedure
\end{algorithmic}
\end{algorithm}

\newpage
\section*{Appendix C: Using cycles of detailed balance kernels}
Recall that we did not require the detailed balance kernels $d_t$ for $t \in \{1\ldots T\}$ to be ergodic. In particular, each $d_t$ can update only a subset of the random variables in $(\theta, e_{1:t}) \in \mathcal{X}_t$. We now show that Algorithm~\ref{alg:smc_experiments} can be used without modification when the kernels $k_t$ for $t \in \{2\ldots T-1\}$ utilize instead \emph{cycles} of detailed balance kernels each targeting the same distribution $p_{t-1}(\theta, e_{1:t-1})$, provided the corresponding kernels $\ell_t$ sample from the same cycle \emph{in reverse order}. Note that the cycle of detailed balance kernels may not itself satisfy detailed balance.

For some $1 < r \le s < T$, suppose that $\mathcal{X}_{r-1} = \mathcal{X}_r = \cdots = \mathcal{X}_s$, and $\tilde{p}_{r-1} = \tilde{p}_r = \cdots = \tilde{p}_s$ (meaning the target distributions do not change from $t=r-1$ through $t=s$). Suppose $k_t(x'; x) = d_{t-1}(x'; x)$ and $\ell_t(x'; x) = d_{t-1}(x'; x)$ for $x, x' \in \mathcal{X}_{r-1}$ for $t \in \{r\ldots s\}$ where $d_{t-1}$ is a detailed balance kernel targeting $\tilde{p}_{r-1}$, for $r \le t \le s$. Then for $r \le t \le s$ the weights are:
\begin{align*}
w_t(x_{t-1}, x_t) = \frac{\tilde{p}_{r-1}(x_t)}{\tilde{p}_{r-1}(x_{t-1})} \frac{d_{t-1}(x_{t-1}; x_t)}{d_{t-1}(x_t; x_{t-1})}
= \frac{\tilde{p}_{r-1}(x_t)}{\tilde{p}_{r-1}(x_{t-1})} \frac{\tilde{p}_{r-1}(x_{t-1})}{\tilde{p}_{r-1}(x_t)} = 1
\end{align*}

Consider modifying the \textproc{simulate} and \textproc{regenerate} procedures to replace $a^i_{t-1} \sim \mbox{Categorical}(\textproc{normalize}(\mathbf{w}_{t-1}))$ with $a^i_{t-1} \gets i$ for $r+1 \le t \le s+1$, and modifying \textproc{rand-ancestry} (used by \textproc{regenerate}) to replace $I_t \sim \mbox{Uniform}(1\ldots N)$ with $I_t \gets I_{t-1}$ for $r+1 \le t \le s+1$. The joint probability for \textproc{simulate} is then divided by the probability of the excluded random choices, which is $1/N^{N(s-r+1)}$ since each weight $w^i_t$ is deterministically $1$ for $r \le t \le s$.
The joint probability for \textproc{regenerate} is then divided by the probability of the excluded random choices, which is $1/N^{(N-1)(s-r+1)} \cdot 1/N^{s-r+1} = 1/N^{N(s-r+1)}$.
The weight expression is therefore unchanged, but can be simplified (because weights $w^i_t$ for $r \le t \le s$ are deterministically $1$) to:
\begin{align*}
\frac{p(u, z)}{q(u;z)}
= \frac{1}{w_{T+1}^1\left(\prod_{\substack{t=1}}^{r-1} \frac{1}{N} \sum_{j=1}^N w^j_t \right) \cdot \left( \prod_{t=s+1}^T  \frac{1}{N} \sum_{j=1}^N w^j_t\right)}
\end{align*}
In a concise implementation of these modified procedures, the steps $r$ through $s+1$ are collapsed into one step, with the cycle of detailed balance kernels in $(k_r = d_{r-1}, \ldots, k_s = d_{s-1}, k_{s+1})$ taking the role of a single $k_t$ in \textproc{simulate} and \textproc{regenerate} and the reverse cycle in $(\ell_{s+1}, \ell_s = d_{s-1}, \ldots, \ell_r = d_{r-1})$ taking the role of the corresponding $\ell_t$ in \textproc{regenerate}.

To see that $k_{T+1}$ and $\ell_{T+1}$ can also be replaced with cycles of detailed balance kernels, consider introducing new random variables $z_1, \ldots, z_R \in \mathcal{X}^T$ for some $R \ge 1$ into both \textproc{simulate} and \textproc{regenerate} as follows: In the joint probability expression $p(u, z)$ replace  $k_{T+1}(z; x_T^{I_t})$ with:
\[
k_{T+1}(z_1; x_T^{I_T}) \prod_{r=2}^R k_{T+r}(z_r; z_{r-1}) k_{T+R+1}(z; z_R)
\]
Let $k_{T+1}, \ldots, k_{T+R+1}$ be detailed balance kernels targeting $\tilde{p}_T$. This corresponds to applying a sequence of detailed balance kernels immediately prior to returning the output $z$ in \textproc{simulate}.
Similarly, in the joint probability expression $q(u;z)$, replace  $\ell_{T+1}(x_T^{I_T}; z)$ with:
\[
k_{T+1}(x_T^{I_T}; z_1) \prod_{r=2}^R k_{T+r}(z_{r-1}; z_r) k_{T+R+1}(z_R; z)
\]
This corresponds to applying the same cycle of detailed balance kernels to the input of \textproc{regenerate} that were applied before the output of \textproc{simulate}, but in the reverse order. The new log weight is then adjusted by a factor of:
\begin{align*}
\frac{k_{T+1}(z_1; x_T^{I_T}) \prod_{r=2}^R k_{T+r}(z_r; z_{r-1}) k_{T+R+1}(z; z_R)}{k_{T+1}(x_T^{I_T}; z_1) \prod_{r=2}^R k_{T+r}(z_{r-1}; z_r) k_{T+R+1}(z_R; z)} \frac{\ell_{T+1}(x_T^{I_T}; z)}{k_{T+1}(z; x_T^{I_T})}\\
= \frac{\tilde{p}_T(z_1)}{\tilde{p}_T(x_T^{I_T})} \prod_{r=2}^R \frac{\tilde{p}_T(z_r)}{\tilde{p}_T(z_{r-1})} \frac{\tilde{p}_T(z)}{\tilde{p}_T(z_R)} \frac{\tilde{p}_T(x_T^{I_T})}{\tilde{p}_T(z)}
= 1
\end{align*}
Therefore, $k_{T+1}$ and $\ell_{T+1}$ can be replaced with a cycle of detailed balance kernels targeting $p_T$ and the reversed cycle, respectively, without modifying the expression for the returned log-weight value, which is still computed using $w_{T+1}^1 = w_{T+1}(x_T^{I_T}, z)$, and does not depend on the intermediate values $z_1,\ldots,z_R$.

\end{document}